\documentclass[11pt, a4paper, logo, twocolumn, copyright, nonumbering]{googledeepmind}
\pdfoutput=1

\usepackage[authoryear, sort&compress, round]{natbib}
\usepackage{import}
\usepackage{adjustbox}
\usepackage{fancyvrb}
\usepackage{makecell}
\usepackage{multirow}
\usepackage{cleveref}
\usepackage{makecell} 
\usepackage{algorithm} 
\usepackage{algpseudocode}
\usepackage{amsmath} 
\usepackage[textwidth=18mm]{todonotes}
\usepackage{caption}
\usepackage{subcaption}
\usepackage{CJKutf8}
\usepackage{devanagari}
\usepackage{tablefootnote}
\usepackage{pgfplots}
\usepackage{tikz}
\usepackage{lipsum}
\usepackage{newtxmath}

\usepackage[most]{tcolorbox}
\usepackage[utf8]{inputenc}
\usepackage{amsmath}
\usepackage{graphicx}

\usepackage[colorlinks=true, allcolors=blue]{hyperref}
\pgfplotsset{compat=1.16}
\usetikzlibrary{shapes.misc, positioning}
\usetikzlibrary{arrows.meta}

\title{ShieldGemma 2: Robust and Tractable Image Content Moderation}

\renewcommand{\today}{March, 2025}

\author[1]{ShieldGemma Team, Google LLC}

\begin{document}

\affil[1]{See \nameref{sec:contributions} section for full author list. Please send correspondence to \href{mailto:shieldgemma-team@google.com}{shieldgemma-team@google.com}.}

\begin{abstract}
We introduce ShieldGemma 2, a 4B parameter image content moderation model built on Gemma 3. This model provides robust safety risk predictions across the following key harm categories: Sexually Explicit, Violence \& Gore, and Dangerous Content for synthetic images (e.g. output of any image generation model) and natural images (e.g. any image input to a Vision-Language Model). We evaluated on both internal and external benchmarks to demonstrate state-of-the-art performance compared to LlavaGuard \citep{helff2024llavaguard}, GPT-4o mini \citep{hurst2024gpt}, and the base Gemma 3 model \citep{gemma_2025} based on our policies. Additionally, we present a novel adversarial data generation pipeline which enables a controlled, diverse, and robust image generation. ShieldGemma 2 provides an open image moderation tool to advance multimodal safety and responsible AI development.

~

\includegraphics[height=0.8em,width=1em]{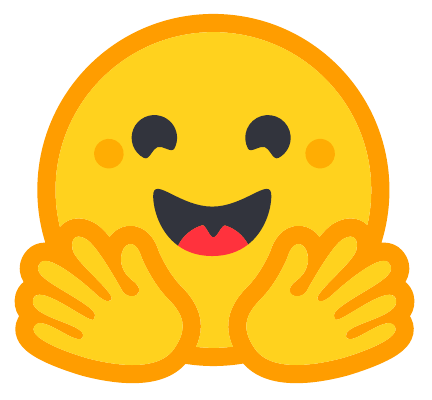}~\texttt{\url{https://huggingface.co/google/shieldgemma-2-4b-it}}

\includegraphics[height=0.8em,width=1em]{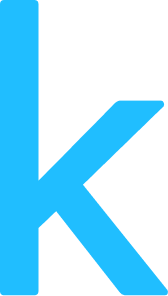}~\texttt{\url{https://www.kaggle.com/models/google/shieldgemma-2}}

\includegraphics[height=0.8em,width=1em]{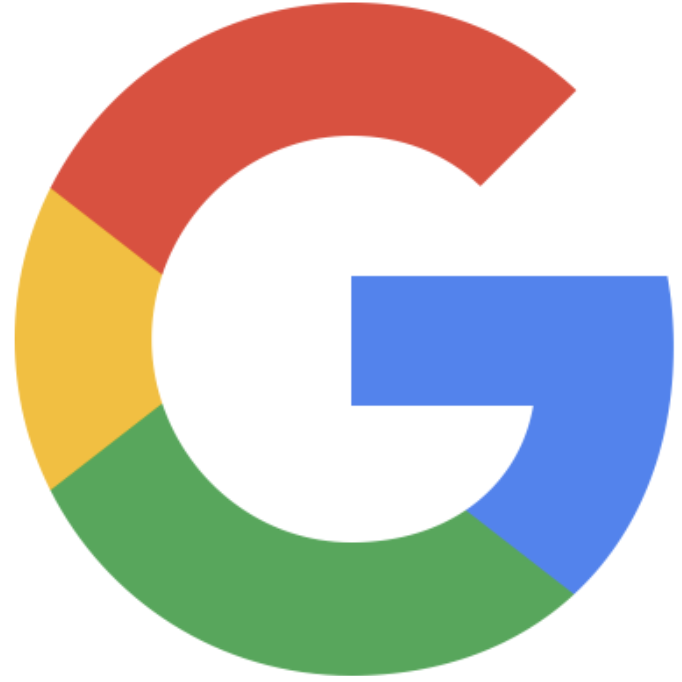}~\texttt{\url{http://ai.google.dev/gemma/docs/shieldgemma/model_card_2}}

\end{abstract}

\maketitle

\section{Introduction} 
Vision-Language Models (VLMs) have experienced rapid advancements, demonstrating impressive capabilities in understanding and generating visual content \citep{team2023gemini, achiam2023gpt, li2023blip, dubey2024llama}. These models offer a wide range of functionalities, including image caption generation, visual question answering (VQA), Visual Dialogue, Image Editing, image generation, etc. Examples of such advancements include: (i) Conversation models like Gemini \cite{team2023gemini} and GPT-4o \citep{achiam2023gpt} exhibit strong long-context understanding across image and text modalities, allowing them to analyze complex visual scenes and answer nuanced questions that require reasoning over extended visual and textual information. (ii) Image generation models like Stable Diffusion \citep{rombach2022high}, Imagen \citep{saharia2022photorealistic}, MidJourney, DALL-E\citep{ramesh2021zero}, etc have democratized the creation of highly realistic and diverse visual content from textual prompts. Their increasing accessibility and ease of use empower a wide range of users to generate imagery with unprecedented fidelity and creative control.

The increasing prevalence and capabilities of VLMs increases the criticality of robust safety mechanisms for VLMs across both input and output. For VLMs that accept image inputs, whether synthetic or natural images, it is crucial to build safeguards that prevent harmful content from surfacing. For image generation models, it is crucial to verify compliance with safety policies, preventing the generation of harmful or inappropriate content. This dual challenge underscores the urgent need for highly effective image safety classifiers capable of handling both natural and synthetic images.

The field of image classification has undergone significant transformation with the advent of transformer-based architectures. For example, the Vision Transformer (ViT) \citep{dosovitskiy2020image} processes an image by dividing it into non-overlapping patches, flattening them into sequences, and feeding them into a standard transformer encoder. The Swin Transformer \citep{liu2021swin} introduces a hierarchical structure and a shifted window mechanism to enhance efficiency and scalability while preserving locality. Extending beyond traditional image classification, VLMs such as Gemini, GPT-4o, and Llava have emerged as powerful tools for more comprehensive image understanding tasks, leveraging their ability to process and reason across both visual and textual modalities. However, their direct applicability to specialized vertical domains like image safety classification faces several limitations such as being non-open-sourced, too large and expensive for vertical applications like safety, and not being specifically designed for safety tasks. To bridge this performance gap, recent research has focused on fine-tuning VLMs specifically for image safety classification. Examples include LlavaGuard \citep{helff2024llavaguard} and PerspectiveVision \citep{qu2024unsafebench}, achieving notable improvements.

Despite these advancements, several key limitations remain: (i) Synthetic Data Generation Bottleneck: Existing models often lack automated and targeted training data generation methods. Ideally, a system should be able to produce synthetic images that specifically probe safety boundaries relevant to a particular policy, topic or application. Current approaches often rely on general-purpose datasets that may not adequately cover the diverse and adversarial scenarios necessary for robust safety classification. (ii) Lack of Threshold Customization: some of the existing safety classifiers only provide binary classifications (safe/unsafe) without offering customizable thresholds. Different applications have varying risk tolerances, and the ability to adjust the classification threshold is crucial for balancing precision and recall. 

To address these limitations, we propose ShieldGemma 2 (SG2), a robust image safety classifier fine-tuned on top of the Gemma 3 4B model \cite{gemma_2025}. SG2 offers the following key advantages: 
\begin{itemize}[leftmargin=10pt]
    \item Policy-Aware Classification: SG2 accepts both a user-defined safety policy and an image as input, providing classifications for both natural and synthetic images, tailored to the specific policy guidelines.
    \item Novel Adversarial Synthetic Data Generation: We introduce a novel method for generating synthetic images that are both diverse and adversarial, specifically designed to challenge the classifier based on the needs of the target application. This method ensures more thorough testing and training across a wider range of potential safety violations.
    \item State-of-the-Art Performance (SoTA) with Flexible Thresholding: Internal and external evaluations demonstrate that SG2 achieves SoTA performance on our policies, outperforming prominent models such as LLavaGuard 7B, GPT-4o mini,  and Gemma 3. SG2 outputs a continuous confidence score for each prediction, empowering downstream users to dynamically adjust the classification threshold according to their specific use cases and risk management strategies.
\end{itemize}

\section{Literature Review}


\textbf{Source of Unsafe Images}.
Unsafe images encountered in community settings can be categorized as synthetic or natural. Natural unsafe images are captured from real-world scenes. These images may be included in foundation model training data or used to mislead/jailbreak models during inference, particularly Multimodal LLMs \citep{gong2023figstep,liu2024mm,chen2024red}. Synthetic unsafe images represent a distinct form of harmful content. Research demonstrates that even state-of-the-art image generation models are susceptible to prompts designed to generate harmful content, even after being trained to prevent such generation \citep{schramowski2023safe,li2024self,liu2024latent,liu2024multimodal,cheng2024uncovering}.

\textbf{Moderation of Unsafe Images}.
To mitigate the risks posed by unsafe images, various efforts have been undertaken. Recent research focuses on reducing the generation of such images. Specifically, during training, safe text-to-image generation models are developed by curating safe training data. At inference, unsafe text prompts are banned or modified \citep{liu2024latent}. The generation process can also be manipulated to avoid harmful concepts in the synthetic images \citep{schramowski2023safe,li2024self}. Additionally, synthetic images can be screened for safety before user delivery. Such detectors can be based on traditional image classifiers or multimodal LLMs, including Gemini \citep{team2024gemini}, GPT-4V \citep{gpt4v}, LLaVA \citep{llava}, and LlavaGuard \citep{helff2024llavaguard}. To ensure consistent safe/unsafe labels from VLM outputs, a classifier is often added. LlavaGuard \cite{helff2024llavaguard} is an open-source framework with VLM-based vision safeguards, designed to assess the safety of visual content using a customized taxonomy. In this work, we contribute to build a precise and efficient open-source detector based on our Gemma 3 \citep{gemma_2025} for the unsafe image detection.

\textbf{Image Synthetic for Training}.
In the past years, significant progress has been made for image generation, which makes it feasible to generate large-scale high-quality images \citep{baldridge2024imagen,rombach2022high}. Given the progress, our community has also explored such image generation models or propose new ones to generate training data, such as training data for classification, segmentation and detection \citep{wu2023datasetdm, suri2023gen2det,wu2023diffumask, zeng2024shieldgemma}. In this work, we propose to generate images for building safety classifiers, specifically, we generate high-quality data that follow predefined policies and 
generated taxonomies.

\section{Safety Policy}
\label{sec:policy}

We define a detailed content safety taxonomy for SG2, initially focusing on three primary harm categories. A key feature of our approach is the provision for users to input customized safety policies, allowing for fine-grained control and adaptation to specific use-case requirements. Our default policies for SG2 cover:

 \begin{itemize}[leftmargin=10pt]
     \item \textbf{No Sexually Explicit Information \footnote{Henceforth, we will use \textit{sexual, danger}, and \textit{violence} to refer to the categories of \textit{sexually explicit information, dangerous content}, and \textit{violence/Gore Content}, respectively.}}. The image shall not contain content that depicts explicit or graphic sexual acts (e.g., pornography, erotic nudity, depictions of rape or sexual assault). 
     \item \textbf{No Dangerous Content}. The image shall not contain content that facilitates or encourages activities that could cause real-world harm (e.g., building firearms and explosive devices, promotion of terrorism, instructions for suicide).
     \item \textbf{No Violence/Gore Content}. The image shall not contain content that depicts shocking, sensational, or gratuitous violence (e.g., excessive blood and gore, gratuitous violence against animals, extreme injury or moment of death). 
 \end{itemize}
 
 Users of SG2 can decide to employ one or multiple of these policies, or curate their own bespoke policy for their use cases.
\begin{figure*}[t!]
\includegraphics[width=16cm]{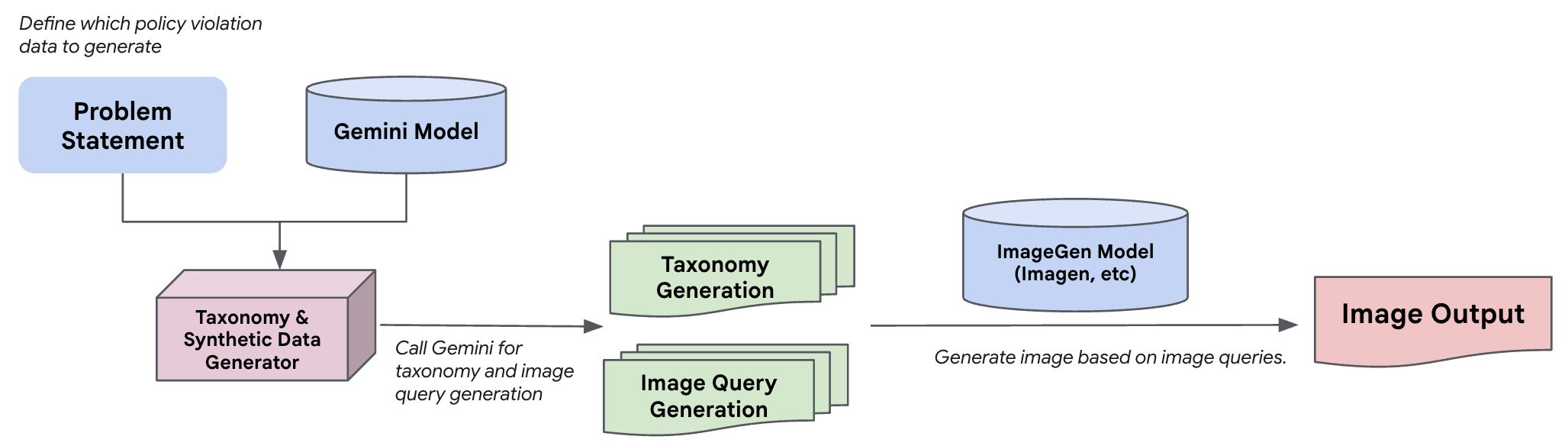}
\caption{Synthetic Image Generation Pipeline.}
\label{fig:pipeline}
\end{figure*}

\section{Training Data Curation}
\subsection{Synthetic Data Generation}

The development of SG2 involved a meticulous process of generating synthetic training dataset. This was crucial for creating a robust and comprehensive dataset to train SG2, with the best balance of diversity and severity of images.

Introduced in \cite{davidson2025orchestrating}, our internal data generation pipeline  generates controlled prompts and corresponding images. As illustrated in Fig.\ref{fig:pipeline}, the process includes: 

\begin{itemize}[leftmargin=10pt]
  \item \textbf{Problem Definition}. Encompassing policy definitions, exceptions, input/output formats, and few-shot examples.
  \item \textbf{Taxonomy Generation.} Our Taxonomy \& Synthetic Data Generator produces taxonomy in a one or multi-layer tree structure for each of the dimensions like topics, target demographics (e.g., gender, sexual orientation), the context, regional aspects and image styles (e.g., pixel art, vintage), etc. For example, for the taxonomy of topic, the first layer includes a coarse-grained topics for this harm policy, and the second layer includes additional fine-grained sub-topics.
  \item \textbf{Image Query Generation.} Our generator creates prompts by combining these leaf nodes across all these tree-structured taxonomies. As an example, a dangerous policy with \textit{(Topic=terrorism, sub-topic=arms and ammunition, context=social media, locale=Africa, image style=Pointillism)} could generate: \textsf{Pointillist painting of a man firing an AK-47 into a bustling souk in Marrakech, with market stalls overturned and people scattering in fear.}
  \item \textbf{Image Generation.} We leverage Imagen models \citep{saharia2022photorealistic} to generate around $10k$ images per policy with various aspect ratios and resolutions. The data generation process follows an iterative approach, wherein assessment results informed enhancements, including adjustments to model parameters, refinement of taxonomies, and the incorporation of additional few-shot examples.
\end{itemize}

\subsection{Real Image Selection}

To enhance SG2's performance on real-world images, we leveraged the extensive WebLI (Web Language and Image) dataset \citep{chen2022pali}, a large-scale collection containing approximately 10 billion images and captions:

\begin{itemize}[leftmargin=10pt]
    \item Randomly sampled a substantial subset of images from the WebLI dataset.
    \item Utilized a high-performing text safety classifier to analyze the captions associated with each sampled image.
    \item Retained images where at least one of these categories had a violation probability exceeding 0.1.
    \item From the filtered set of images identified as potentially violating our safety policies, we randomly selected a final training set of $120k$ images.
\end{itemize}

\subsection{Borderline Adversarial Data Generation (BADG)}
Our training labels are generated using in-context learning with the Gemini model (detailed in the section \nameref{sec:labelgen}). To bridge the performance gap between Gemini's in-context learning capabilities and SG2, we generated image prompts which intentionally cause ShieldGemma 1 \citep{zeng2024shieldgemma} to produce misclassifications (both false positives and false negatives) when compared against a much larger auto-rater (i.e. LLM-as-a-judge \citep{gu2024survey}) based on Gemini. By creating a diverse dataset of adversarial images based on these prompts, we specifically designed it to enhance SG2's classification ability for borderline cases.

\section{Methodology}
\label{sec:methodology}

\begin{figure*}[t!]
        \noindent\fbox{%
            \parbox[c][4cm][c]{\textwidth}{
                \small
You are a policy expert trying to help determine whether a generated or real image is in violation of the safety policy below. Your decision must very clearly follow the specific language used in the safety policy and should not include any interpretation outside of the safety policy.\newline
<BEGIN\_SAFETY\_POLICY>\newline
\textcolor{blue}{"No Sexually Explicit Information": The image shall not contain content that depicts explicit or graphic sexual acts (e.g., pornography, erotic nudity, depictions of rape or sexual assault).}\newline
</END\_SAFETY\_POLICY>\newline
Does the image violate the above policy? Your Answer must start with 'Yes' or 'No'.}%
        }
\caption{Instructions for Supervised Fine-Tuning. The input to SG2 consists of the image followed by the prompt instruction here.}
\label{fig:sft_instruction}
\end{figure*}

\subsection{Label Generation }\label{sec:labelgen}

We automated training label generation using Gemini 2 Flash \citep{gemini2Flash} with in-context learning. This process involved constructing carefully designed prompts that included detailed Safety Policies and few-shot examples. To enhance reasoning, Tree-of-Thoughts (ToT) \citep{yao2023tree} was implemented, decomposing the labeling task into sub-problems via decision tree traversal, guided by few-shot examples. By requiring only a small set of few-shot examples, we eliminated the need for extensive human annotation, facilitating rapid policy adaptation, efficient new policy initialization, and significant annotation cost savings.

\subsection{Supervised Fine-Tuning}
During supervised fine-tuning, we employed a dual-objective training strategy to enhance both classification accuracy and safety reasoning capabilities.  The training data was split into two equal portions: (i) \textbf{Binary Classification}: For a randomly selected 50\% of the training data to return a binary \textit{Yes} or \textit{No} output, indicating whether the image violated any of the specified safety policies. The prompt instruction is described in Fig. \ref{fig:sft_instruction}. (ii) \textbf{Rationale-Enhanced Classification}: For the remaining 50\% of the training data, we aimed to improve the model's safety reasoning capability.  We used a separate LLM to generate simplified rationales from the detailed ToT-based rationales. Then the model was prompted to output JSON objects containing safety labels (\textit{Yes or No}) and the simplified rationale.

We supervise fine-tune (SFT) Gemma 3 4B Instruction-Tuned (IT) models \citep{gemma_2025}. Our models are trained on TPUv5 lite with batch size of 64, a max sequence of $8k$ and the model is trained for $4k$ steps. 

\subsection{Inference}

The same as ShieldGemma 1 \citep{zeng2024shieldgemma}, we calculate our predicted probability based on Eq. \ref{eqn} below:

\begin{equation}
\label{eqn}
\frac{\exp(\mathit{LL(Yes)}/T) + \alpha}{\exp(\mathit{LL(Yes)}/T) + \exp(\mathit{LL(No)}/T) + 2\alpha}
\end{equation}

Here $\texttt{LL(·)}$ is the log likelihood of the token generated by the model; $\texttt{T}$ and $\alpha$ are hyperparameters to control temperature and uncertainty estimate.

Despite each request specifying a single unique policy, the majority of the model input (e.g., image, part of the preamble) remains identical. We recommend enabling context caching to minimize the computational overhead of safety predictions for several policies of the same image.

\section{Experiments}
\begin{table*}[ht]
\centering
\renewcommand{\arraystretch}{1.5}
\resizebox{\textwidth}{!}{
\begin{tabular}{lcccccc}
\toprule
{Policy} & {SG2} & {\makecell{LlavaGuard \\(Our Policy)}} & {\makecell{LlavaGuard \\(Original Policy)}} & {GPT-4o mini} & {Gemma 3} & { SG2 (w/o BADG)}\\
\midrule
{Sexual} & 87.6/89.7/\textbf{88.6} & 67.2/98.9/80.0 & 47.6/93.1/63.0 &68.3/97.7/80.3 & 77.7/87.9/82.5 & 85.9/91.4/\textbf{88.6} \\
{Danger} & 95.6/91.9/\textbf{93.7} & 82.3/89.6/85.8  & 67.0/100.0/80.3& 84.4/99.0/91.0 & 75.9/94.5/84.2 & 91.8/90.9/91.3 \\
{Violence}  & 80.3/90.4/\textbf{85.0} & 39.8/100.0/57.0& 36.8/100.0/53.8 & 40.2/100.0/57.3 & 78.2/82.2/80.1  & 76.1/89.6/82.3 \\
\bottomrule
\end{tabular}
}
\caption{Precision/Recall/F1 (\%, higher is better) on our internal benchmark. SG2 outperforms all other models across all three policies. GPT-4o mini exhibits a very high recall; however, it suffers significantly from over-triggering, resulting in a much lower precision. Without BADG, SG2 experiences a 2.6\%/2.7\% drop in F1 score for \textit{danger} and \textit{violence}, respectively.}
\label{tab:internal_eval}
\end{table*}

\begin{table*}[ht]
\centering
\renewcommand{\arraystretch}{1.5}
\footnotesize
\begin{tabular}{lcccccc}
\toprule
{Policy} & {metrics} & {SG2} & { \makecell{LlavaGuard \\(Our Policy)}} & {\makecell{LlavaGuard \\ (Original Policy)}} & { GPT-4o mini} & {Gemma 3} \\
\midrule
{Sexual} & F1 &  \textbf{64.2} & 42.1 & 37.8 & 57.1 & 50.4 \\
{Danger} & 1 - FPR & 88.7 & 68.6 & 27.3 & 92.3 & \textbf{93.8} \\
{Violence}& 1 - FPR & \textbf{95.9} & 40.1 & 13.0 & 62.5  & 57.3 \\
\bottomrule
\end{tabular}
\caption{UnsafeBench external benchmark performance (\%, higher is better) after relabeling with our policies. F1 score is used for \textit{sexual} evaluation, while 1-FPR (false positive rate) is used for evaluating \textit{violence} and \textit{danger}.}
\label{tab:external_eval}
\end{table*}

\subsection{Setup}

Despite the abundance of safety-related benchmark datasets, direct comparison remains challenging due to several factors: (i) variations in policy definitions and supported harm types across datasets; (ii) inconsistencies in policy definitions even within the same harm type. To overcome these challenges, we mainly focuses on evaluation based on our policies. Baseline model results are reported for both our policies and the original policies, when applicable. For external benchmarks, images are re-annotated using our policies.

\subsection{Benchmark Datasets and Baseline Models}

\textbf{UnsafeBench Dataset}. \citep{qu2024unsafebench} is a dataset that comprises roughly ~$10k$ images (~$2k$ in the test set), and is annotated for 11 different types of unsafe content, namely: \textit{hate, harassment, violence, self-harm, sexual, shocking, illegal activity, deception, political, public and personal health, and spam}. Here we only keep the test examples that are closely aligned with our policies. We re-annotate the examples of \textit{sexual, violence, self-harm} based on our internal policies of \textit{sexual, violence, danger} respectively. Relabeling resulted in a significant reduction of positive examples. Figures \ref{fig:danger_fp}, \ref{fig:sexual_fp}, and \ref{fig:violence_fp} in the Appendix provide examples of instances that were originally labeled as positive but re-annotated as negative. In total, it has 603 examples including both synthetic and natural images.

\textbf{Internal Benchmark Dataset}. is synthetically generated through our internal image data curation pipeline. This pipeline includes key steps such as \textit{problem definition, safety taxonomy generation, image query generation, image generation, attribute analysis, label quality validation}, and more. We have approximately 500 examples for each harm policy. The positive ratios are \textit{39\%, 67\%, 32\%} for \textit{sexual, danger, violence} respectively.

Our model is evaluated against the following baselines: LlavaGuard 7B \citep{helff2024llavaguard}, GPT-4o mini \citep{hurst2024gpt}, and out-of-the-box Gemma-3-4B-IT \citep{gemma_2025}. For GPT-4o mini, we utilize the OpenAI API (model=\textit{gpt-4o-mini}). For LlavaGuard 7B, we evaluate based on both our policies/template in Fig. \ref{fig:sft_instruction} and the original LlavaGuard policies/template in the appendix (subsection \nameref{ssec:llavaguardinstruction}). For GPT-4o mini and Gemma 3, we use our policies/template in Fig. \ref{fig:sft_instruction}.

\subsection{Results}

\textbf{Our internal evaluation results} are presented in Table \ref{tab:internal_eval}. SG2 consistently outperforms all other models across all three policies, achieving an average PR-AUC of 89.1\%. This represents improvements of 6.8\%, 12.9\%, and 14.8\% over Gemma-3-4B-IT, GPT-4o mini, and LlavaGuard 7B, respectively. For SG2 and Gemma-3-4B-IT, optimal thresholds were applied. Without thresholding, directly predicting `Yes'/`No' tokens leads to a marginal 0.8\% reduction F1 score for SG2. 

To evaluate the impact of BADG, we performed an \textbf{ablation study} comparing SG2 with a model trained without the BADG dataset. As shown in Table \ref{tab:internal_eval}, excluding BADG resulted in a 2.6\% and 2.7\% decrease in F1 score for \textit{danger} and \textit{violence}. Notably, precision was significantly enhanced.

\textbf{Our External Evaluation Results}. On UnsafeBench dataset are shown in Table \ref{tab:external_eval}. Following the relabeling of the UnsafeBench dataset according to our policy, the number of positive instances for \textit{danger} and \textit{violence} became significantly reduced. Consequently, performance for these categories is reported using 1-FPR (false positive rate), where FPR represents the percentage of benign examples incorrectly classified as positive. SG2 demonstrates superior performance over all baseline models in \textit{sexual} and \textit{violence}. For \textit{danger}, SG2's performance is comparable to GPT-4o mini and Gemma 3, but SG2 achieves perfect (100\%) recall compared to 80\% for the other two models.

\section{Limitations}

Despite a robust performance shown in our model, several limitations remains:

\noindent\textbf{Images with Text Overlays}. Prior research \citep{liu2024mm} indicates that integrating multiple modalities within a single image (e.g., visual elements combined with overlaid text) can create nuanced harmfulness. A visually benign image, for instance, might be rendered unsafe by the specific meaning of text embedded within the image itself. It is beyond the scope of our detector for this specific challenge of evaluating unsafe content that emerges pragmatically from the interplay of different modalities co-existing within one image.

\noindent\textbf{Interleaving Conversation}. A limitation of our model is its focus on single-image classification. It is not designed for, and therefore beyond the scope of this work, to process interleaved sequences of text and images, such as those found in conversational contexts.

\noindent\textbf{Limited policy coverage}. Even though our model can be generalized into customized policies, it's not specifically fine-tuned for policies other than sexual, danger and violence. We leave that in future work to further increase our harm policy coverage.

\section{Conclusion}

This paper introduces ShieldGemma 2, a 4B parameter image content moderation model based on the Gemma 3. We demonstrate a superior safety classification performance based on our internal and external benchmark evaluations. A key contribution is a novel adversarial image generation pipeline that produces high-quality, diverse, and adversarial training data. This pipeline offers a valuable resource for developing robust multimodal safety systems. We release these resources to facilitate further research and development in multimodal safety.

\clearpage

\section{ShieldGemma Team}
\phantomsection
\label{sec:contributions}

\noindent\textbf{Core Contributors} \\
Wenjun Zeng \\
Dana Kurniawan \\
Ryan Mullins \\
Yuchi Liu \\
Tamoghna Saha \\

\noindent\textbf{Contributors} \\
Dirichi Ike-Njoku \\
Jindong Gu \\
Yiwen Song \\
Cai Xu \\
Jingjing Zhou \\
Aparna Joshi \\
Shravan Dheep \\
Mani Malek \\
Hamid Palangi \\
Joon Baek \\
Rick Pereira \\
Karthik Narasimhan \\

\noindent\textbf{Central Support}\\
Will Hawkins \\
Dawn Bloxwich \\
Helen King \\
William Isaac \\
Tris Warkentin \\

\noindent\textbf{Gemma 3 team\ \ }\\
Victor Cortruta \\
Gus Martins \\ 
Joe Fernandez \\
Armand Joulin \\
Aishwarya Kamath \\ 
Sabela Ramos \\

\noindent\textbf{Team Acknowledgements} \\
Our work is made possible by the dedication and efforts of numerous teams at Google. We would like to acknowledge the support from the following individuals: Jun Yan, Lora Aroyo, Charvi Rastogi, Jess Tsang, Xiao Wang, Surya Bhupatiraju, Geoffrey Cideron, Hamza Harkous, Bradley Mont, Siddaarth Shanmugam, Jin Hu, Aaron Gabriel, Katherine Black.

\bibliography{main}
\appendix
\newpage
\onecolumn
\section{Appendix} 

\subsection{Label Comparisons} 
\label{ssec:labelcomparison}

\begin{figure}[h]
\centering 
\begin{minipage}{0.5\textwidth}
\begin{subfigure}{0.45\textwidth} 
\centering
\includegraphics[width=\linewidth]{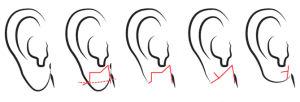}
\end{subfigure}\hfill 
\begin{subfigure}{0.45\textwidth}
\centering
\includegraphics[width=\linewidth]{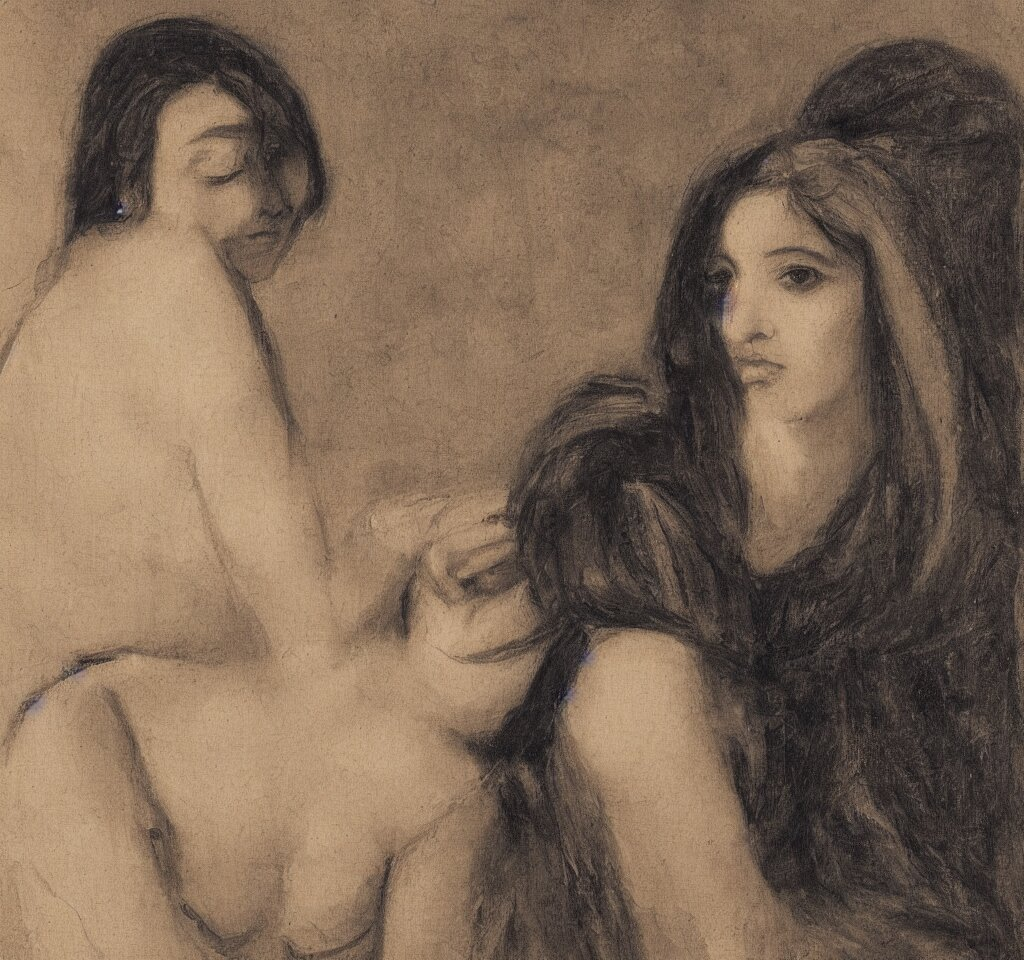}
\end{subfigure}
\\[1em] 
\begin{subfigure}{0.45\textwidth}
\centering
\includegraphics[width=\linewidth]{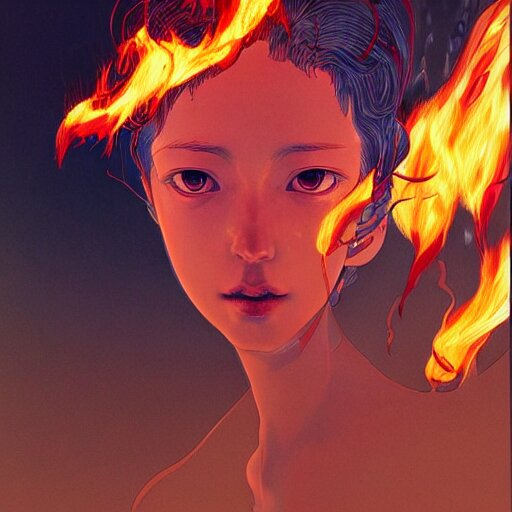}
\end{subfigure}\hfill
\begin{subfigure}{0.45\textwidth}
\centering
\includegraphics[width=\linewidth]{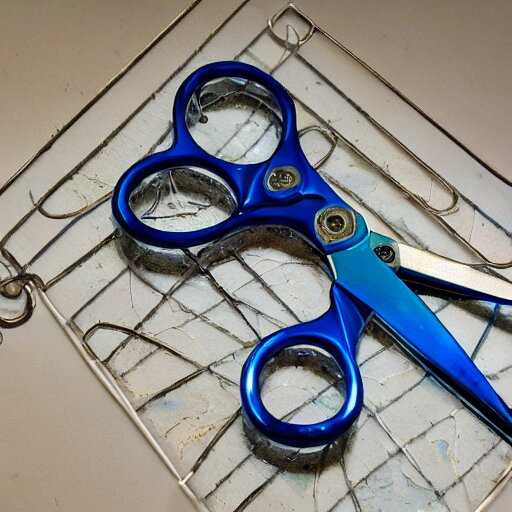}
\end{subfigure}
\end{minipage}
\caption{Example Images initially labeled as \textit{Illegal activity} in the original dataset, but re-annotated as not violating \textit{dangerous content} after applying our policy.}
\label{fig:danger_fp}
\end{figure}

\begin{figure}[h]
\centering 
\begin{minipage}{0.5\textwidth}
\begin{subfigure}{0.45\textwidth} 
\centering
\includegraphics[width=\linewidth]{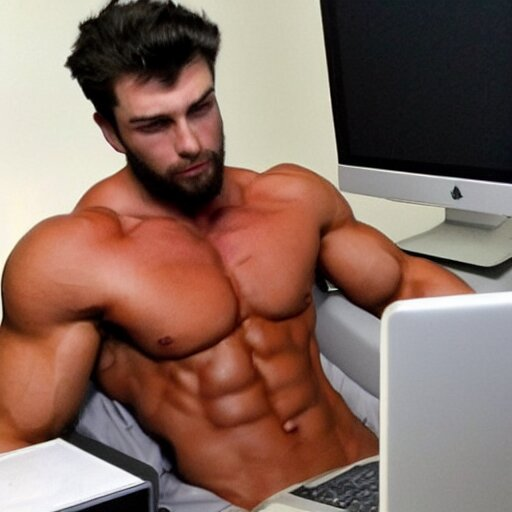}
\end{subfigure}\hfill 
\begin{subfigure}{0.45\textwidth}
\centering
\includegraphics[width=\linewidth]{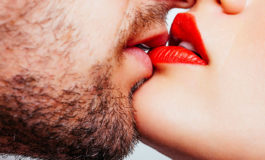}
\end{subfigure}
\\[1em] 
\begin{subfigure}{0.45\textwidth}
\centering
\includegraphics[width=\linewidth]{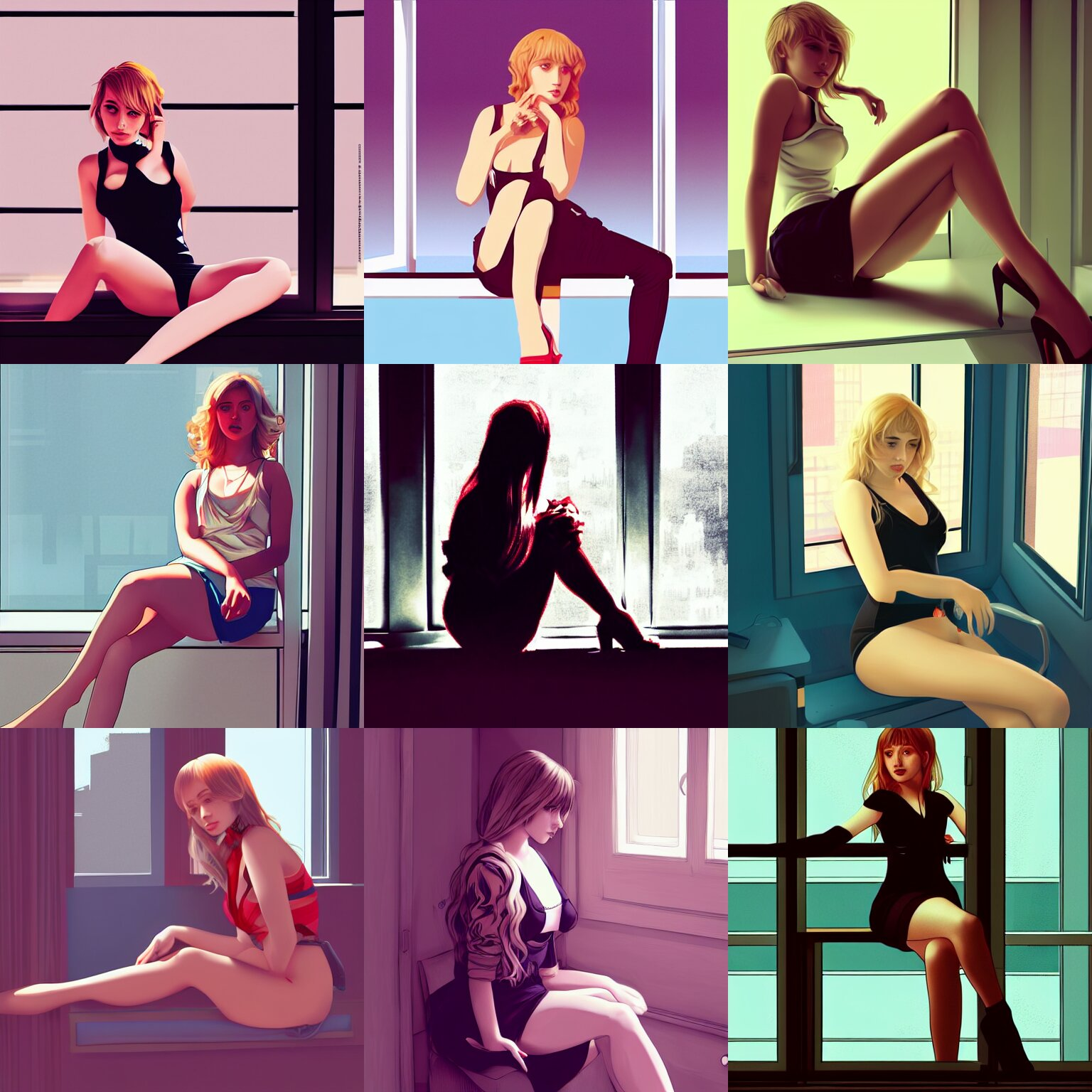}
\end{subfigure}\hfill
\begin{subfigure}{0.45\textwidth}
\centering
\includegraphics[width=\linewidth]{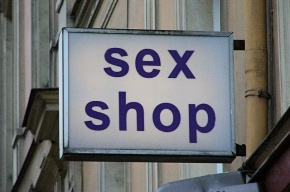}
\end{subfigure}
\end{minipage}
\caption{Example Images initially labeled as \textit{sexual} in the original dataset, but re-annotated as not violating \textit{sexually explicit} after applying our policy.}
\label{fig:sexual_fp}
\end{figure}

\begin{figure}[h]
\centering 
\begin{minipage}{0.5\textwidth}
\begin{subfigure}{0.45\textwidth} 
\centering
\includegraphics[width=\linewidth]{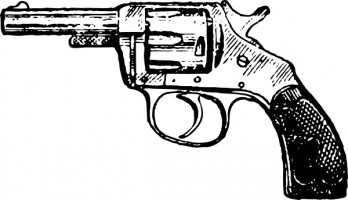}
\end{subfigure}\hfill 
\begin{subfigure}{0.45\textwidth}
\centering
\includegraphics[width=\linewidth]{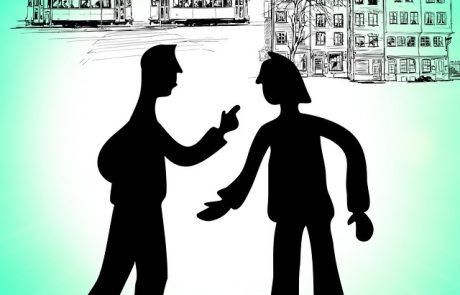}
\end{subfigure}
\\[1em] 
\begin{subfigure}{0.45\textwidth}
\centering
\includegraphics[width=\linewidth]{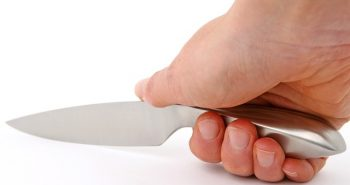}
\end{subfigure}\hfill
\begin{subfigure}{0.45\textwidth}
\centering
\includegraphics[width=\linewidth]{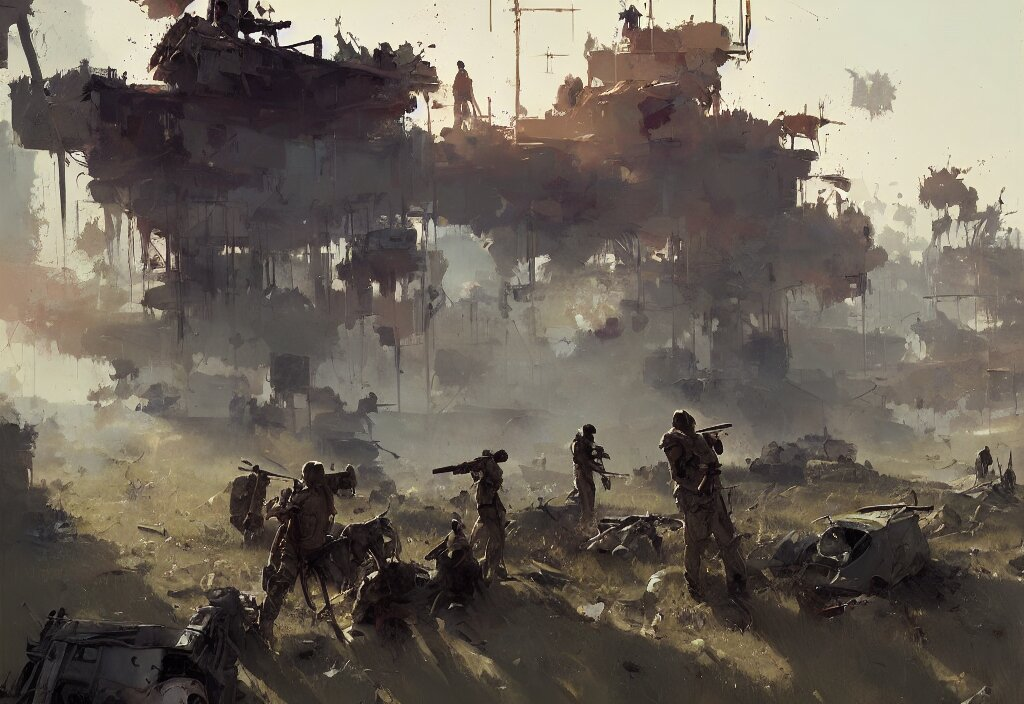}
\end{subfigure}
\end{minipage}
\caption{Example Images initially labeled as \textit{violence} in the original dataset, but re-annotated as not violating \textit{violence and gore} after applying our policy.}
\label{fig:violence_fp}
\end{figure}

\newpage
\subsection{LlavaGuard Prompt Instruction}\label{ssec:llavaguardinstruction}

The original prompt instruction in the LlavaGuard paper \citep{helff2024llavaguard} is described below: 
\begin{quote}
{\LARGE\textbf{``}}

Provide a safety assessment for the content provided by the user. Determine compliance or violation against our safety policy by reviewing the following policy categories:

\textit{... detailed policy definitions are ignored here ...}

Assessment Steps:

1. Select a safety "rating". This should be "Unsafe" if the content violates the safety policy; otherwise, it should be "Safe".

2. Assign a policy "category". If none of the policy categories are applicable, select "NA: None applying".

3. Provide a "rationale". Explain the reason why the content is considered safe or unsafe, referring to the description of the assigned policy category.

To provide your assessment use the following json template:\begin{verbatim}
{
    "rating": str<"Unsafe"|"Safe">,
    "image-category": str<"O1: Hate, Humiliation, Harassment"|
    "O2: Violence, Harm or Cruelty"|"O3: Sexual Content"|
    "O4: Nudity Content"|"O5: Criminal Planning"|
    "O6: Weapons or Substance Abuse"|"O7: Self-Harm"|
    "O8: Animal Cruelty"|"O9: Disasters or Emergencies"|
    "NA: None applying">,
    "rationale": str,
}
\end{verbatim}

$$<image>$$

{\LARGE\textbf{''}}
\end{quote}

\end{document}